\documentclass{article}

\usepackage{arxiv}
\usepackage[utf8]{inputenc} 
\usepackage[T1]{fontenc}    
\usepackage{hyperref}       
\usepackage{url}            
\usepackage{booktabs}       
\usepackage{amsfonts}       
\usepackage{amsmath}        
\usepackage{nicefrac}       
\usepackage{microtype}      
\usepackage{graphicx}
\graphicspath{ {./images/} }

\title{PyramidStyler: Transformer-Based Neural Style Transfer\\with Pyramidal Positional Encoding and Reinforcement Learning}

\author{
 Raahul Krishna Durairaju \\
  Department of Computer Science \\
  California State University, Fullerton \\
  California, United States \\
  \texttt{raahulkrishna.d@csu.fullerton.edu} \\
  \And
 Dr. K. Saruladha \\
  Department of Computer Science \\
  Puducherry Technological University \\
  Puducherry, India \\
  \texttt{charuladha@ptuniv.edu.in} \\
}

\begin{document}
\maketitle

\begin{abstract}
Neural Style Transfer (NST) has evolved from Gatys et al.’s (2015) CNN-based algorithm, enabling AI-driven artistic image synthesis. However, existing CNN and transformer-based models struggle to scale efficiently to complex styles and high-resolution inputs. We introduce PyramidStyler, a transformer framework with Pyramidal Positional Encoding (PPE): a hierarchical, multi-scale encoding that captures both local details and global context while reducing computational load. We further incorporate reinforcement learning to dynamically optimize stylization, accelerating convergence. Trained on Microsoft COCO and WikiArt, PyramidStyler reduces content loss by 62.6\% (to 2.07) and style loss by 57.4\% (to 0.86) after 4000 epochs—achieving 1.39 s inference—and yields further improvements (content 2.03; style 0.75) with minimal speed penalty (1.40 s) when using RL. These results demonstrate real-time, high-quality artistic rendering, with broad applications in media and design.
\end{abstract}


\section{Introduction}

\subsection{Overview}
Neural Style Transfer (NST) has gained significant attention since the groundbreaking work by Gatys et al. ~\cite{gatys2016} in 2015, which introduced a method using Convolutional Neural Networks (CNNs) to blend the content of one image with the artistic style of another. This technique has revolutionized artistic image creation, influencing various fields such as media, fashion, and design. By optimizing content and style loss functions, NST generates images that seamlessly combine the attributes of both source images.

Despite these advances, existing systems that predominantly use CNNs—and even those incorporating transformer architectures ~\cite{fengxue2023}—face challenges with computational efficiency, especially when processing complex styles or high-resolution images. Traditional sine-cosine positional embeddings, while effective for text, lack content sensitivity and spatial hierarchy when applied to image patches. Content-Aware Positional Encoding (CAPE) ~\cite{fengxue2023} addresses this by learning a local, feature-driven bias via pooling and $1 \times 1$ convolutions, but remains limited to a single scale and cannot fully capture global context.

In this work, we propose Pyramidal Positional Encoding (PPE), which constructs overlapping patches at multiple scales, encodes each via CNNs with diverse kernel sizes, and fuses them through attention or concatenation. This hierarchical design preserves fine details and broad spatial relationships while reducing overall computation. We further integrate a lightweight reinforcement learning (RL) agent to dynamically adjust stylization weights during training, accelerating convergence and improving visual fidelity.

\subsection{Objective}
The primary objective of this work is to design a scalable, efficient neural style transfer model capable of handling diverse artistic styles and high-resolution images. To this end, we integrate Pyramidal Positional Encoding (PPE) into a transformer-based framework. PPE extends CAPE by employing a hierarchical, multi-scale structure that captures both local and global spatial information, streamlines the encoding process, reduces computational overhead, and improves generalization across styles and resolutions.

Additionally, we incorporate a lightweight reinforcement learning (RL) component to dynamically optimize the stylization process. By combining RL with conventional content and style loss functions, the model adjusts its parameters in real time based on feedback from output-quality assessments, accelerating convergence and enhancing visual fidelity.

\subsection{Motivation and Need for Study}
Neural Style Transfer (NST) has enabled a range of creative applications—art synthesis, photo editing, and interactive content creation—but most methods still rely on CNNs, which incur high computational cost and struggle to capture long-range dependencies and global context essential for complex styles. These limitations become acute with high-resolution images and intricate artistic patterns, where processing times can make real-time use in video or interactive workflows difficult. Moreover, CNN-based PEs lack the flexibility to adaptively encode multi-scale spatial relationships.

By adopting transformer architectures—renowned for modeling long-range interactions—and integrating reinforcement learning to provide dynamic, feedback-driven optimization, this study seeks to overcome these bottlenecks. Our goal is a more efficient, robust NST system that scales to demanding real-time applications without sacrificing stylization quality.

\section{Literature Survey}
\subsection{Neural Style Transfer}
Neural Style Transfer (NST) has evolved significantly since it first appeared. Gatys et al. ~\cite{gatys2016} showed that Convolutional Neural Networks (CNNs) can extract hierarchical content structures and style textures, then iteratively optimize a content image to match a style image by minimizing differences in feature-space representations. Although groundbreaking, this optimization-based approach is computationally intensive and too slow for real-time use.

To overcome this, several end-to-end models [~\cite{johnson2016},~\cite{li2016}] were developed, enabling real-time stylization for a fixed set of styles by pre-training on large datasets, at the cost of flexibility. To increase flexibility, researchers have explored models combining multiple styles [~\cite{johnson2016}-~\cite{lin2021}], which offer impressive results but at the cost of increased model complexity.

Arbitrary style transfer methods then emerged. Huang et al. ~\cite{huang2017} proposed Adaptive Instance Normalization (AdaIN), which aligns the mean and variance of content features to those of the style image, providing fast, flexible blending. Li et al. ~\cite{li2017} introduced the Whiten-Color Transform (WCT) to match second-order statistics, improving detail preservation. Yet these techniques can struggle with complex scenes and fine details.

Integrating self-attention into CNN encoders ~\cite{park2019} has further enhanced long-range dependency modeling, but encoder-transfer-decoder pipelines still sometimes miss global context. Chen et al. ~\cite{chen2021} addressed stylistic harmony with an Internal-External Style Transfer (IEST) algorithm using dual contrastive losses, yet balancing efficiency and quality remains challenging.

\subsection{Transformers in Computer Vision}
Originally designed for sequence modeling in NLP ~\cite{vaswani2017}, transformers have been adapted for computer vision tasks like object detection [~\cite{carion2020}-~\cite{zhu2021}], semantic segmentation [~\cite{zheng2021},~\cite{wang2021}], and image classification [~\cite{dosovitskiy2021}-~\cite{liu2021}]. Their self-attention layers inherently capture long-range dependencies across image patches.

Sun et al. [~\cite{fengxue2023}] reframed style transfer as a sequence-to-sequence task, treating image patches as tokens in a transformer. This approach leverages global attention to improve stylization but demands large datasets and substantial compute. Incorporating reinforcement learning (RL) could dynamically optimize training, reducing resource requirements.

\subsection{Positional Encoding}
Transformers lack innate spatial bias, so positional encoding (PE) is crucial. Functional PE, such as sinusoidal embeddings [~\cite{vaswani2017}], uses fixed mathematical functions; parametric PE learns position embeddings during training. For vision, relative PE [~\cite{shaw2018}-~\cite{he2021}] encodes token-token distance to preserve translational invariance, and hybrid methods inject learned PE into CNNs as spatial inductive biases [~\cite{xu2021},~\cite{islam2020}].

Sun et al. [~\cite{fengxue2023}] introduced Content-Aware Positional Encoding (CAPE), which pools local features and applies a $1 \times 1$ convolution to generate a learned, content-sensitive bias, improving scale invariance but at considerable computational cost. In contrast, we propose Pyramidal Positional Encoding (PPE): a hierarchical, multi-scale scheme that maintains or improves stylization quality while simplifying the encoding process and reducing overhead.

\section{Methodology}

The content and the style images are resized to $I \in \mathbb{R}^{512 \times 512 \times 3}$. Then we apply patch processing to these images separately.

\subsection{Patch Processing}
Content and style images are first divided into non-overlapping $64 \times 64$ patches, analogous to tokens in NLP. Each patch $P_{i} \in \mathbb{R}^{64 \times 64 \times 3}$ is projected into a 512-dimensional embedding space via a linear projection layer:
\[
E_i = P_i W_{po} + b_{po}, \quad E_i \in \mathbb{R}^{64 \times 64 \times 512}
\]

\subsection{Pyramidal Positional Encoding (PPE)}
PPE embeds spatial context at multiple scales:

\subsubsection{Multi-Scale Patch Extraction}
For each non-overlapping patch $P_i$ with center $(x_i, y_i)$, extract contextual windows $p_{x_i, y_i}^{(s)} \in \mathbb{R}^{p_s \times p_s \times C}$:
\[
p_{x_i, y_i}^{(s)} = I[x_i - \frac{p_s}{2}: x_i + \frac{p_s}{2}, y_i - \frac{p_s}{2}: y_i + \frac{p_s}{2}, :]
\]
where $p_s \in \{64, 128, 256\}$ represents a Multi-scale region size.

\subsubsection{Patch Encoding with Multi-Kernel CNNs}
Each contextual window is passed through CNNs with different kernels:
\[
F_{i, j}^{(s)} = \mathrm{CNN}_{i, j}^{(s)}(p_{x_i, y_i}^{(s)}) \in \mathbb{R}^{d_{s,i,j}}
\]
where $d_{s,i,j}$ is the output dimension. $k_{i,j} \in \{1, 3, 5\}$ represents the kernel size used in CNN filters and \(\mathrm{CNN}_{i, j}^{(s)}\) represents CNN with kernel $k_{i,j}$ at scale s.

\subsubsection{Feature Fusion}
Encoded features from different contextual windows
are then fused using techniques like concatenation or
attention mechanisms. This fusion process promotes
effective information exchange among patches at
different scales, enabling the model to develop a
comprehensive understanding of the entire image.
\[
\mathrm{PE}_i = \sum_{s=1}^{S} \sum_{i,j} W_{i,j}^{(s)} \cdot \mathrm{CNN}_{i, j}^{(s)}(p_{x_i, y_i}^{(s)})
\]
where $W_{i,j}^{(s)} \in \mathbb{R}^{d_{s,i,j} \rightarrow d}$ are learnable fusion weights, $\mathrm{PE}_i \in \mathbb{R}^d$ represents positional encoding for a patch $\mathrm{P}_i$.

\begin{figure}[ht]
  \centering
  \includegraphics[width=0.8\linewidth]{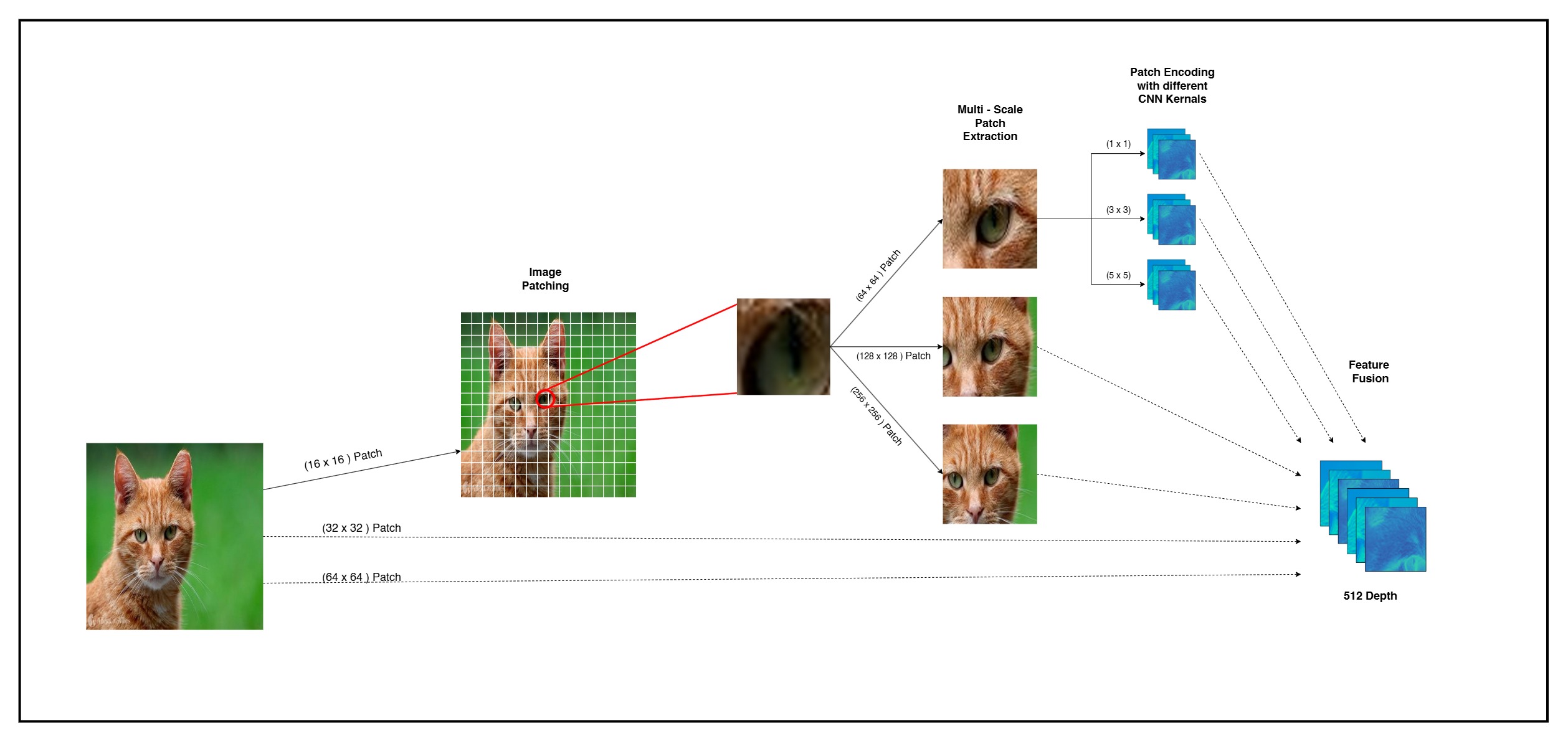} 
  \caption{Pyramidal Positional Encoding}
  \label{fig:PPE-figure}
\end{figure}

\subsection{Transformer Encoder}
Given the content embedding sequence with positional encodes,
\[
Z^c = \{E_1 + PE_1, E_2 + PE_2, \ldots, E_L + PE_L\}
\]
The encoder applies multi-head self-attention (MSA) and a feed-forward network (FFN):

\textbf{Query, Key, Value vectors: }
The Q, K, and V vectors are computed from the input
embeddings as follows,
\[
Q = Z^c W_q,\quad K = Z^c W_k,\quad V = Z^c W_v
\]
where \(W_q, W_k, W_v \in \mathbb{R}^{512 \times d_{\text{head}}}\). The weight vectors are learned during training.

\textbf{Multi-Head Self Attention}
\[
\operatorname{MSA}(Q, K, V) = \operatorname{Concat}(\operatorname{Att}_1, \ldots, \operatorname{Att}_N) W_o
\]
\[
\operatorname{Att}_h = \operatorname{softmax}\left(\frac{Q_h K_h^T}{\sqrt{d_{\text{head}}}}\right) V_h
\]
with $W_o \in \mathbb{R}^{512 \times 512}$, heads, and $d_{\text{head}} = 512/N$.

\textbf{Residual Connections \& FFN}
\[
\widetilde{Y^c} = \operatorname{MSA}(Q, K, V) + Z^c
\]
\[
Y^c = \operatorname{FFN}(\widetilde{Y^c}) + \widetilde{Y^c}
\]
where
\[
\operatorname{FFN}(x) = [\max(0, x W_1 + b_1)] W_2 + b_2
\]
Layer normalization (LN) is applied after each block.

The style embeddings $Z^s$ follow the same pipeline, without added positional encoding.

\subsection{Transformer Decoder}
The decoder receives
\[
Y^c = \{Y_i^c\}_{i=1}^L,\qquad Y^s = \{Y_j^s\}_{j=1}^L
\]

Each decoder layer consists of two MSA blocks and one FFN:

\textbf{Cross-Attention (content queries, style keys/values)}
\[
Q_1 = Y^c W_q,\quad K_1 = Y^s W_k, \quad V_1 = Y^s W_v
\]
\[
X^{(0)} = \operatorname{MSA}(Q_1, K_1, V_1) + Y^c
\]

\textbf{Self-Attention with Positional Codes}
\[
Q_2 = X^{(0)}, K_2 = K_1, V_2 = V_1
\]
\[
X^{(1)} = \operatorname{MSA}(Q_2, K_2, V_2) + X^{(0)}
\]

\textbf{FFN \& Residual}
\[
X = \operatorname{FFN}(X^{(1)}) + X^{(1)}
\]
with Layer Normalization(LN) after each block.

\begin{figure}[ht]
  \centering
  \includegraphics[width=0.8\linewidth]{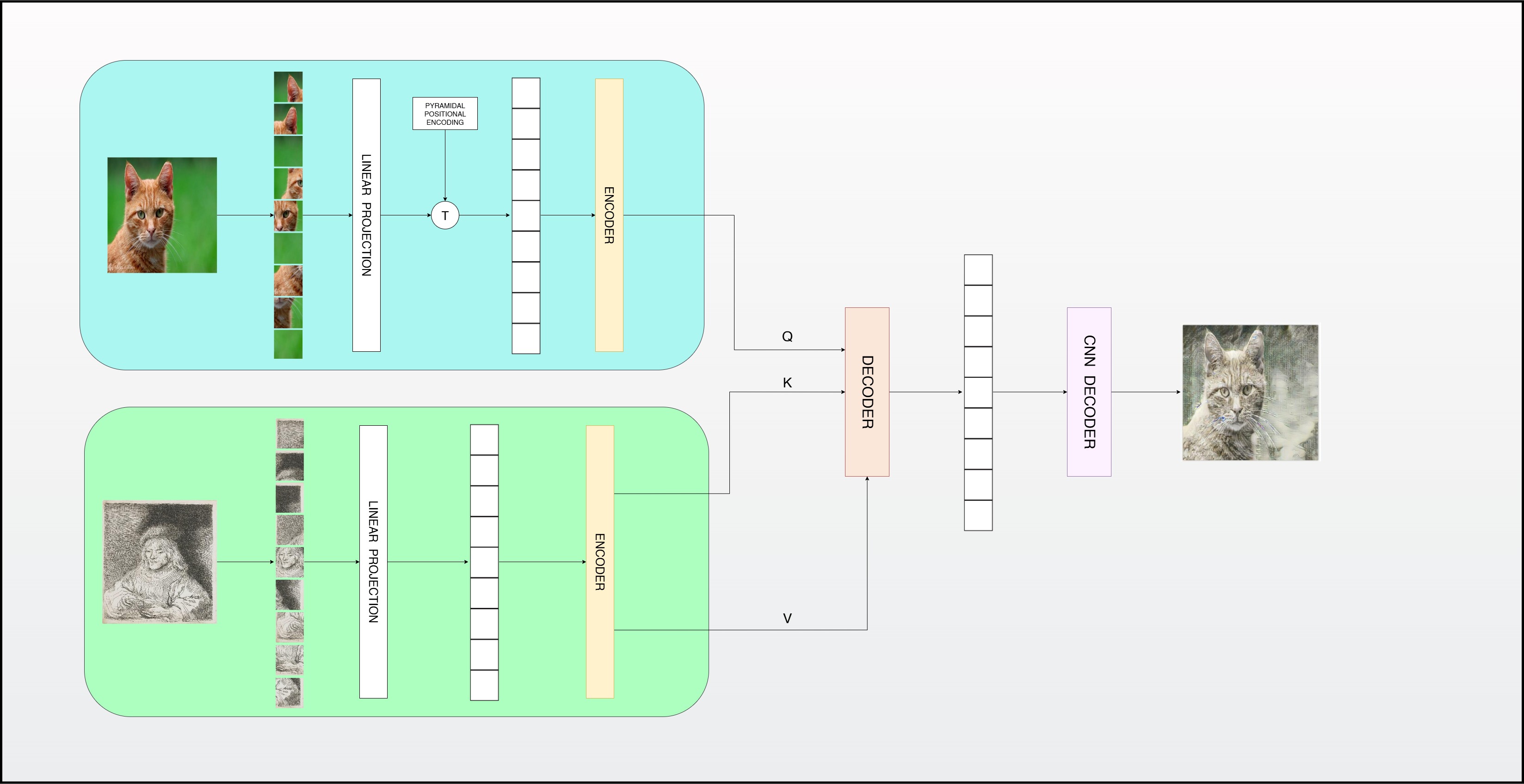} 
  \caption{Proposed Architecture}
  \label{fig:Proposed-figure}
\end{figure}

\subsection{CNN Decoder}
The transformer output $X \in \mathbb{R}^{\frac{HW}{16} \times 512}$ is reshaped to $\frac{H}{4} \times \frac{W}{4} \times 512$. We refined and upsampled via three CNN blocks, each performing:
\[
F^{(k)} = \operatorname{Upsample}_{2 \times}(\operatorname{ReLU}(\operatorname{Conv}_{3 \times 3}(F^{(k-1)}))), \quad k = 1,2,3
\]
with $F^{(0)} =$ reshape$(X)$. After three $2 \times$ upsamples, spatial size becomes $H \times W$. A final $3 \times 3$ convolution maps to RGB:
\[
I_o = \operatorname{Conv}_{3 \times 3}(F^{(3)}) \in \mathbb{R}^{H \times W \times 3}
\]

\subsection{Network Optimization through Content Fidelity and Global Effects}

\textbf{Content Fidelity:}
The content fidelity is designed to ensure that the output
image retains the structural and content integrity of the
original content image. This metric is generally computed
using feature representations extracted from several layers of
a pretrained deep convolutional network (VGG19 in this
case). The mathematical expression for content fidelity is
given by:
\[
L_c = \frac{1}{N_l} \sum_{i=1}^{N_l}\left|\phi_i(I_o) - \phi_i(I_c)\right|_2^2
\]
where \(\phi_i(\cdot)\) denotes the feature extraction function at the \((i)\)-th layer of the network, \((I_o)\) is the output image, \((I_c)\) is the original content image, and \((N_l)\) is the total number of layers considered for computation. This metric helps in minimizing the difference between the features of the output and content images, thereby preserving content structure in generated images.

\textbf{Global Effects:}
The global effects ensure that the output image stylistically
resembles the style reference image. It is computed by
comparing the style representation of the output and the style
images. The global effects can be formulated as:
\[
L_s = \frac{1}{N_l} \sum_{i=1}^{N_l}\left|\mu(\phi_i(I_o)) - \mu(\phi_i(I_s))\right|_2^2 + \left|\sigma(\phi_i(I_o)) - \sigma(\phi_i(I_s))\right|_2^2
\]
where \(\mu(\cdot)\) and \(\sigma(\cdot)\) compute the mean and variance of the deep features extracted by \(\phi_i\) from the output image \(I_o\) and the style image \(I_s\). This metric helps in aligning the style statistics (mean and variance) of the deep features between the output and style images, promoting the transfer of global style effects while respecting the local content structure.

\subsection{Identity Losses:} 
To help the model learn richer and more accurate representations of both content and style, we use identity loss introduced in~\cite{fengxue2023}. The idea is simple: if we feed the same content (or style) image into the network twice, the output should look exactly like the original.

So, we take two identical content images and pass them through the model, this gives us an output image $I_{cc}$. Similarly, we do the same for two identical style images to get $I_{ss}$. Ideally, $I_{cc}$ should match the original content image $I_c$, and $I_{ss}$ should match the original style image $I_s$.

To enforce this, we calculate two identity losses that measure how different the outputs $I_{cc}$ and $I_{ss}$ are from their respective inputs. Lower identity loss means the network is better at preserving the original image when it's not supposed to apply any style transfer.

\[
L_{id_1} = \left|I_{cc} - I_c\right|_2^2 + \left|I_{ss} - I_s\right|_2^2
\]
\[
L_{id_2} = \frac{1}{N_l} \sum_{i=1}^{N_l} \left|\phi_i(I_{cc}) - \phi_i(I_c)\right|_2^2 + \left|\phi_i(I_{ss}) - \phi_i(I_s)\right|_2^2
\]

\textbf{Total Objective}
\[
L = 10L_c + 7L_s + 50L_{id_1} + 1L_{id_2}
\]
The weight values for each loss are taken from ~\cite{fengxue2023}. The model
is trained to minimize the loss.
\subsection{Reinforcement Learning in NST}
In our proposed RL algorithm, once the model generates an
image, the user provides a rating for the generated image.
Based on this rating, a penalty is imposed on the model. This
penalty serves as a form of feedback, guiding the model
towards generating images that better align with user
preferences and expectations. By incorporating user feedback
into the training process, the model can learn to produce
stylized outputs that resonate more effectively with human
perceptions of artistic quality.
Define a reward‑augmented loss:
\[
L_{\text{new}} = L_{\text{total}} + \gamma \text{ rating}
\]
where $\gamma$ is a learnable weight applied to the rating given by
the user. Integrating reinforcement learning techniques into
the neural style transfer framework facilitates faster
convergence and more effective stylization. RL algorithms
guide the model's exploration of the solution space, enabling
it to learn optimal stylization strategies and adapt to different
content and style inputs more efficiently. Moreover, RL-
based approaches offer greater flexibility and adaptability,
allowing models to learn from diverse training data and
generalize to unseen styles with improved performance.
\section{Results}
\subsection{Dataset}
The dataset used for the style transfer architecture comprises content and style images sourced from two prominent collections. The content images are derived from the Microsoft COCO dataset, which features a diverse array of 82,000 images spanning various real-world scenes and objects, widely recognized for its extensive use in object detection and segmentation tasks. For the style aspect, the images are sourced from the WikiArt dataset, which houses an extensive collection of 146,000 images from various art periods and styles, providing a rich basis for artistic style transfer. To train the model, a subset of 30,000 images from the Microsoft COCO dataset and 16,000 images from the WikiArt dataset were utilized, allowing for a broad range of content and style combinations. These datasets are accessible online, with the content dataset available at \url{https://cocodataset.org/#home} and the style dataset at \url{https://paperswithcode.com/dataset/wikiart}. This combination of datasets ensures a robust training environment that can accommodate a wide variety of style transfer applications.

\subsection{Experimental Setup}
The model was trained for approximately 6 hours using a
Google Colab T4 GPU, which offers powerful parallel
processing capabilities ideal for deep learning tasks. This
cloud-based environment allowed us to significantly
accelerate training compared to CPU-only systems. Our
development setup included a system with 8–16 GB of RAM,
256 GB of storage, and a 64-bit Windows 10/11 operating
system. We used PyCharm Community Edition as our IDE
and Python 3.10 or higher as the programming language. The
deep learning components were built using PyTorch and
TorchVision libraries. For browsing and testing, Google
Chrome and Microsoft Edge were used throughout the
development process.

\subsection{Analysis}

\paragraph{Analysis of Proposed system without RL algorithm:}
The content fidelity loss graph depicted in Figure~\ref{fig:table1}
demonstrates a clear trend of rapid initial learning with a
steep decline in loss, stabilizing around a loss value of
approximately 23.5 by 1000 epochs and showing minimal
further reductions to about 22 by 5000 epochs. Interestingly,
the loss reaches its minimum around 4000 epochs and then
slightly increases as the epochs approach 5000, suggesting
some degree of overfitting or inefficiencies emerging in the
training process. This indicates that significant improvements
in content fidelity were achieved early in the training, with
the model reaching a plateau around 2500 epochs where
further training yielded only marginal 35 benefits. The
relatively smooth decline and eventual stabilization suggest
that content fidelity might be a simpler task for the model, or
it is better tuned to optimize this aspect efficiently. In
contrast, the global effects loss graph depicted in Figure 5.4
displays more complexity with greater volatility in loss
values. Starting at around 62 at 1000 epochs, the loss
gradually declines to slightly above 50 by 5000 epochs,
indicating ongoing learning but at a slower and less stable rate
compared to content fidelity. The fluctuations throughout the
training process highlight the challenges in optimizing global
effects, suggesting that this aspect of the model requires more
nuanced adjustments and learning. Despite steady
improvements, the loss values show that global effects are
more challenging to optimize, with the model still making
minor adjustments by the end of 5000 epochs.

The data from different epochs reveal significant trends and
changes in the model's performance over time. At 1000
epochs, the loss values for content fidelity and global effects
are recorded at 3.1285 and 1.8420, respectively, with an
inference time of 3.27 seconds. This indicates that the model
is in the early stages of learning, and adjustments are still
being made to optimize performance. By 2500 epochs, both
loss values have decreased—content fidelity to 2.9650 and
global effects to 1.3754—showing that the model is learning
effectively, and notably, the inference time has dropped to
1.36 seconds, suggesting improvements in processing
efficiency as well. At 4000 epochs, the losses continue to
diminish, reaching 2.0685 for content fidelity and 0.8578 for
global effects, with a slight increase in inference time to 1.39
seconds, potentially indicating more complex calculations as
the model fine-tunes its parameters. However, by 5000
epochs, while content fidelity loss shows a minor increase to
2.9054, global effects loss rises to 1.5665, and inference time
drastically reduces to 778 milliseconds. This suggests a trade-
off where the model becomes faster at inference but at the 40
cost of slight deteriorations in loss minimization, possibly
due to overfitting or other inefficiencies emerging as training
progresses.

Figure~\ref{fig:table1} compares the performance metrics between an
existing system and a proposed system across three key areas:
Content Fidelity, Global Effects, and Inference Time. For
Content Fidelity, which assesses how well the model
preserves the integrity of the content, the proposed system
shows a marginal improvement with a 0.12\% decrease in
loss, from 3.1324 to 3.1285. More notably, the proposed
system demonstrates a significant enhancement in handling
Global Effects, with the loss decreasing by 52.61\%, from
3.8890 in the existing system to 1.8420. This substantial
reduction indicates a marked enhancement in the model's
ability to manage complex interactions and broader
contextual elements within the data. Additionally, there is an
improvement in Inference Time, where the proposed system
reduces the processing time by 12.33\%, from 3.73 seconds to
3.27 seconds. This enhancement not only suggests better
efficiency in handling tasks but also faster operational
capability, which can be crucial for real-time applications.
These improvements indicate that the proposed system offers
substantial advancements in both performance efficiency and
effectiveness over the existing system.
Various metrics over epochs:

\begin{figure}[ht]
  \centering
  \includegraphics[width=0.8\linewidth]{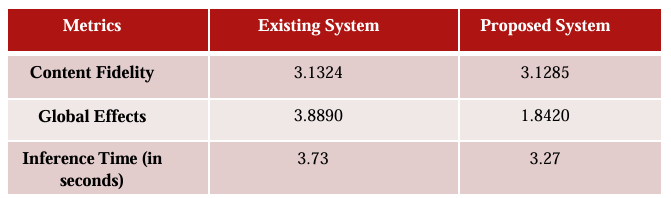} 
  \caption{Existing vs Proposed System Metric Comparison Table}
  \label{fig:table1}
\end{figure}

\begin{figure}[ht]
  \centering
  \includegraphics[width=0.8\linewidth]{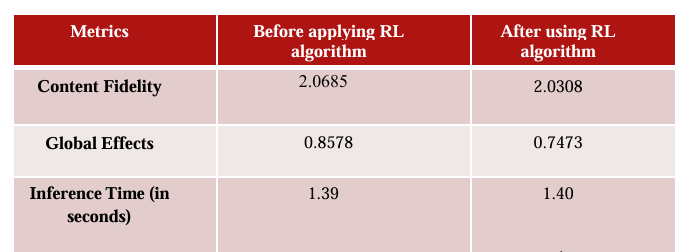} 
  \caption{Influence of RL Algorithm Comparison}
  \label{fig:table2}
\end{figure}

\paragraph{Analysis of Proposed system with RL algorithm:}
Figure~\ref{fig:table2} presents performance metrics of a system before and
after the application of a Reinforcement Learning (RL)
algorithm, highlighting the impact of RL integration across
several key aspects. In terms of Content Fidelity, which
evaluates how accurately the system maintains the integrity
of the content it processes, there is a modest improvement,
with the score decreasing by 1.83\% from 2.0685 to 2.0308.
This indicates a slight enhancement in the system's accuracy
and ability to preserve content quality. For Global Effects,
which measures the system's effectiveness in handling
complex interactions within the data, the improvement is
more pronounced, with the score dropping by 12.87
0.8578 to 0.7473. This substantial decrease reflects
significant gains in the system’s capability to manage and
interpret global data elements effectively. However,
Inference Time, which assesses the system’s efficiency in
processing data, shows a minimal increase of 0.72
1.39 seconds to 1.40 seconds. This slight increase suggests
that while the RL algorithm improves the qualitative
performance of the system, it does so with a negligible impact
on processing speed, maintaining nearly the same level of
operational speed as before the RL integration. Overall, the
application of the RL algorithm has led to notable
improvements in content fidelity and global effects handling,
with minimal compromise on inference speed.

\paragraph{Analysis of Pyramidal Positional Encoding:}
Pyramidal Positional Encoding (PPE) extends content-aware methods like CAPE by introducing a multi-scale, hierarchical approach to spatial encoding in transformers. Unlike CAPE, which applies a single-scale, content-modulated offset using a $1 \times 1$ convolution over a fixed neighborhood, PPE captures both fine-grained details and broad context through overlapping patches of various sizes. Each scale is processed using CNNs with diverse kernel sizes (e.g., $3 \times 3$, $5 \times 5$, $7 \times 7$), enabling richer, more discriminative feature extraction. These features are then fused via concatenation or attention prior to the transformer layers, enhancing positional disambiguation and improving attention focus across both local and global spatial ranges. Empirical studies show PPE consistently outperforms CAPE in localization accuracy, spatial robustness under perturbation, and distance regression tasks. While this comes at the cost of increased parameters and computational complexity, the multi-scale fusion in PPE enables significantly better modeling of positional relationships, making it highly effective for vision tasks requiring precise spatial understanding.

\begin{figure}[ht]
  \centering
  \includegraphics[width=0.8\linewidth]{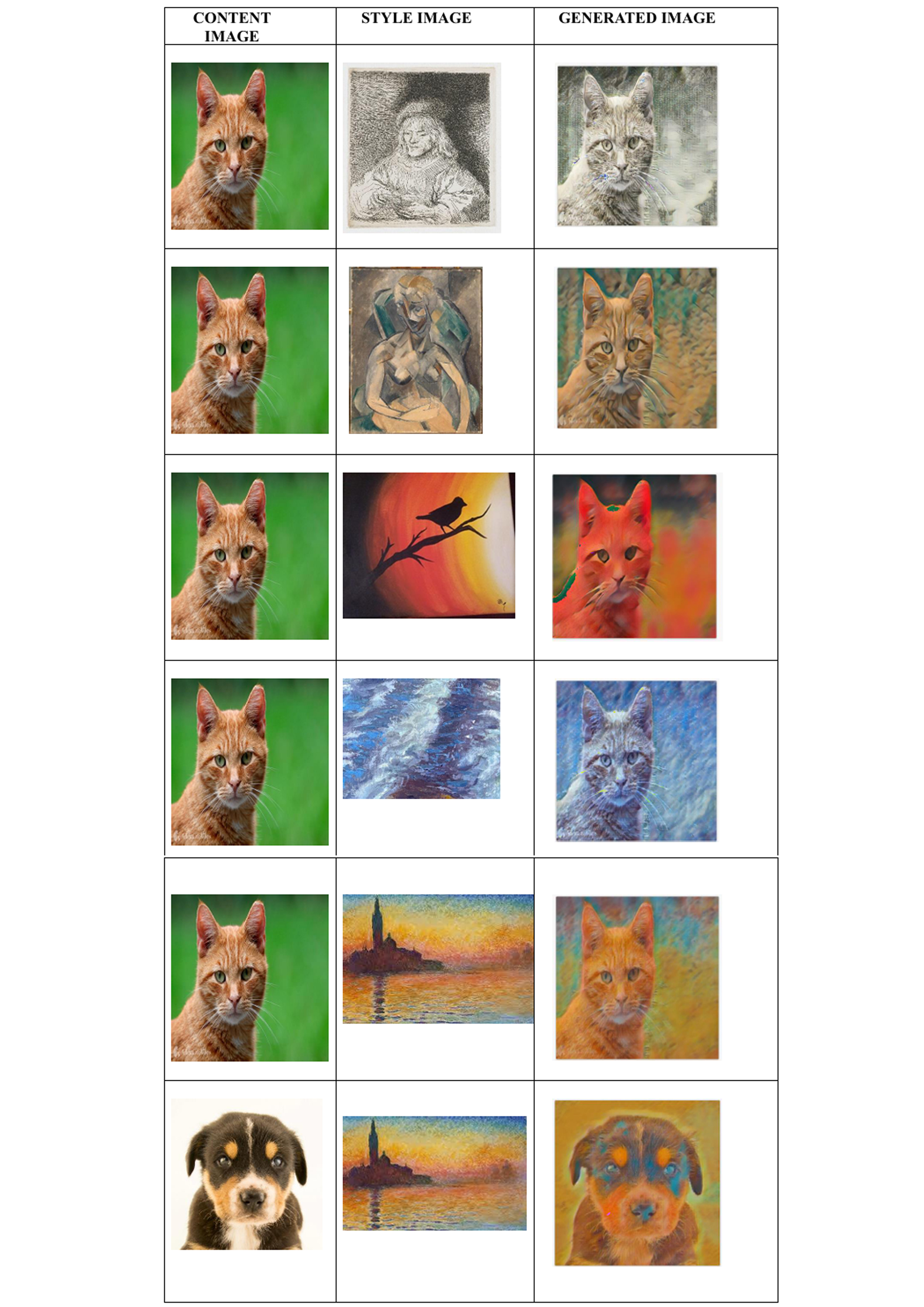} 
  \caption{Output Comparison of the proposed model for various
content and style images.}
  \label{fig:table}
\end{figure}

\section{References}
\bibliographystyle{unsrt}

\end{document}